\documentclass[twoside,11pt]{article}

\usepackage{jmlr2e}

\usepackage{color}

\newcommand*\samethanks[1][\value{footnote}]{\footnotemark[#1]}
\firstpageno{1}

\usepackage{lastpage}
\jmlrheading{22}{2021}{1-\pageref{LastPage}}{3/21; Revised
8/21}{9/21}{21-0343}{Meng Liu, Youzhi Luo, Limei Wang, Yaochen Xie, Hao Yuan, Shurui Gui, Haiyang Yu, Zhao Xu, Jingtun Zhang, Yi Liu, Keqiang Yan, Haoran Liu, Cong Fu, Bora Oztekin, Xuan Zhang, and Shuiwang Ji.}
\ShortHeadings{DIG: A Turnkey Library for Diving into Graph Deep Learning Research}{Liu et al.}

\begin{document}

% \title{DIG: A Turnkey Library for Diving into Graph Neural Network Research}
\title{DIG: A Turnkey Library for Diving into Graph Deep Learning Research}

\author{\name Meng Liu\thanks{These authors contributed equally.} \email mengliu@tamu.edu \vspace{-0.22cm}
       \AND
\name Youzhi Luo\samethanks[1] \email yzluo@tamu.edu \vspace{-0.22cm}
       \AND
\name Limei Wang\samethanks[1] \email limei@tamu.edu \vspace{-0.22cm}
       \AND
\name Yaochen Xie\samethanks[1] \email ethanycx@tamu.edu \vspace{-0.22cm}
       \AND
\name Hao Yuan\samethanks[1] \email hao.yuan@tamu.edu \vspace{-0.22cm}
       \AND       
\name Shurui Gui\samethanks[1] \email shurui.gui@tamu.edu \vspace{-0.22cm}
       \AND   
\name Haiyang Yu\samethanks[1] \email haiyang@tamu.edu \vspace{-0.22cm}
       \AND 
\name Zhao Xu \email zhaoxu@tamu.edu \vspace{-0.22cm}
       \AND
\name Jingtun Zhang \email zjt6791@tamu.edu \vspace{-0.22cm}
       \AND 
\name Yi Liu \email yiliu@tamu.edu \vspace{-0.22cm}
       \AND 
\name Keqiang Yan \email keqiangyan@tamu.edu \vspace{-0.22cm}
       \AND
\name Haoran Liu \email liuhr99@tamu.edu \vspace{-0.22cm}
       \AND  
\name Cong Fu \email congfu@tamu.edu \vspace{-0.22cm}
       \AND
\name Bora Oztekin \email bora@tamu.edu \vspace{-0.22cm}
       \AND 
\name Xuan Zhang \email xuan.zhang@tamu.edu \vspace{-0.22cm}
       \AND
\name Shuiwang Ji \email sji@tamu.edu \vspace{-0.1cm} \\
       \addr Department of Computer Science and Engineering\\
       Texas A\&M University\\
       College Station, TX 77843-3112, USA}

% \editor{Kevin Murphy and Bernhard Sch{\"o}lkopf}

\editor{Alexandre Gramfort}

\maketitle

\begin{abstract}%   <- trailing '%' for backward compatibility of .sty file
Although there exist several libraries for deep learning on graphs, they are aiming at implementing basic operations for graph deep learning. In the research community, implementing and benchmarking various advanced tasks are still painful and time-consuming with existing libraries. To facilitate graph deep learning research, we introduce \textit{DIG: Dive into Graphs}, a turnkey library that provides a unified testbed for higher level, research-oriented graph deep learning tasks. Currently, we consider graph generation, self-supervised learning on graphs, explainability of graph neural networks, and deep learning on 3D graphs. For each direction, we provide unified implementations of data interfaces, common algorithms, and evaluation metrics. Altogether, \textit{DIG} is an extensible, open-source, and turnkey library for researchers to develop new methods and effortlessly compare with common baselines using widely used datasets and evaluation metrics.
Source code is available at \url{https://github.com/divelab/DIG}.
\end{abstract}

\begin{keywords}
  graph deep learning, generation, self-supervised learning, explainability, 3D graphs, Python
\end{keywords}

\section{Introduction}

% General description of graph deep learning. General introduction of existing graph deep learning packages.

Graph deep learning~\citep{bronstein2017geometric,grl_ideb17,wu2020comprehensive,zhou2018graph,battaglia2018relational,hamilton2020graph,ma2020deep} has been drawing increasing attention due to its effectiveness in learning from rich graph data. It has achieved remarkable successes in many domains, such as social networks~\citep{kipf2016semi,velivckovic2017graph,hamilton2017inductive}, drug discovery~\citep{gilmer2017neural,wu2018moleculenet,stokes2020deep,wang2020advanced}, and physical simulation~\citep{battaglia2016interaction,sanchez2020learning}. Several libraries, such as PyG~\citep{Fey/Lenssen/2019}, DGL~\citep{wang2019dgl}, tf\_geometric~\citep{hu2021efficient}, Spektral~\citep{grattarola2020graph}, GraphNet~\citep{battaglia2018relational}, StellarGraph~\citep{StellarGraph}, GraphGallery~\citep{li2021graphgallery}, CogDL~\citep{cen2021cogdl}, and OGB~\citep{hu2020ogb}, have been developed to facilitate deep learning on graphs. However, most existing libraries focus on providing basic components of graph neural networks and mainly consider elementary tasks, such as node classification and graph classification. With these libraries, it still costs a lot of efforts to implement and benchmark algorithms for advanced tasks, such as graph generation.

% Motivation of proving DIG. General introduction of DIG, including the main contents, goal, and key features.
To bridge this gap, we present a Python library \textit{DIG: Dive into Graphs}. We currently consider several research directions in graph deep learning. These are graph generation, self-supervised learning on graphs, explainability of graph neural networks, and deep learning on 3D graphs. For each direction, $\textit{DIG}$ provides unified and extensible implementations of data interfaces, common algorithms, and evaluation metrics. \textit{DIG} naturally endows researchers the convenience of developing their algorithms and conducting empirical comparisons with baselines. Altogether, DIG is an extensible, open-source, and turnkey library for researchers to develop new methods and effortlessly compare with common baselines using widely used datasets and evaluation metrics.

\section{Library Description}\label{sec:library}

% Mention the dependencies. Currently, DIG covers 4 topics and 14 algorithms with data interfaces and evaluation metrics(A figure).

\begin{figure}[t]
    \begin{center}
        \includegraphics[width=\textwidth]{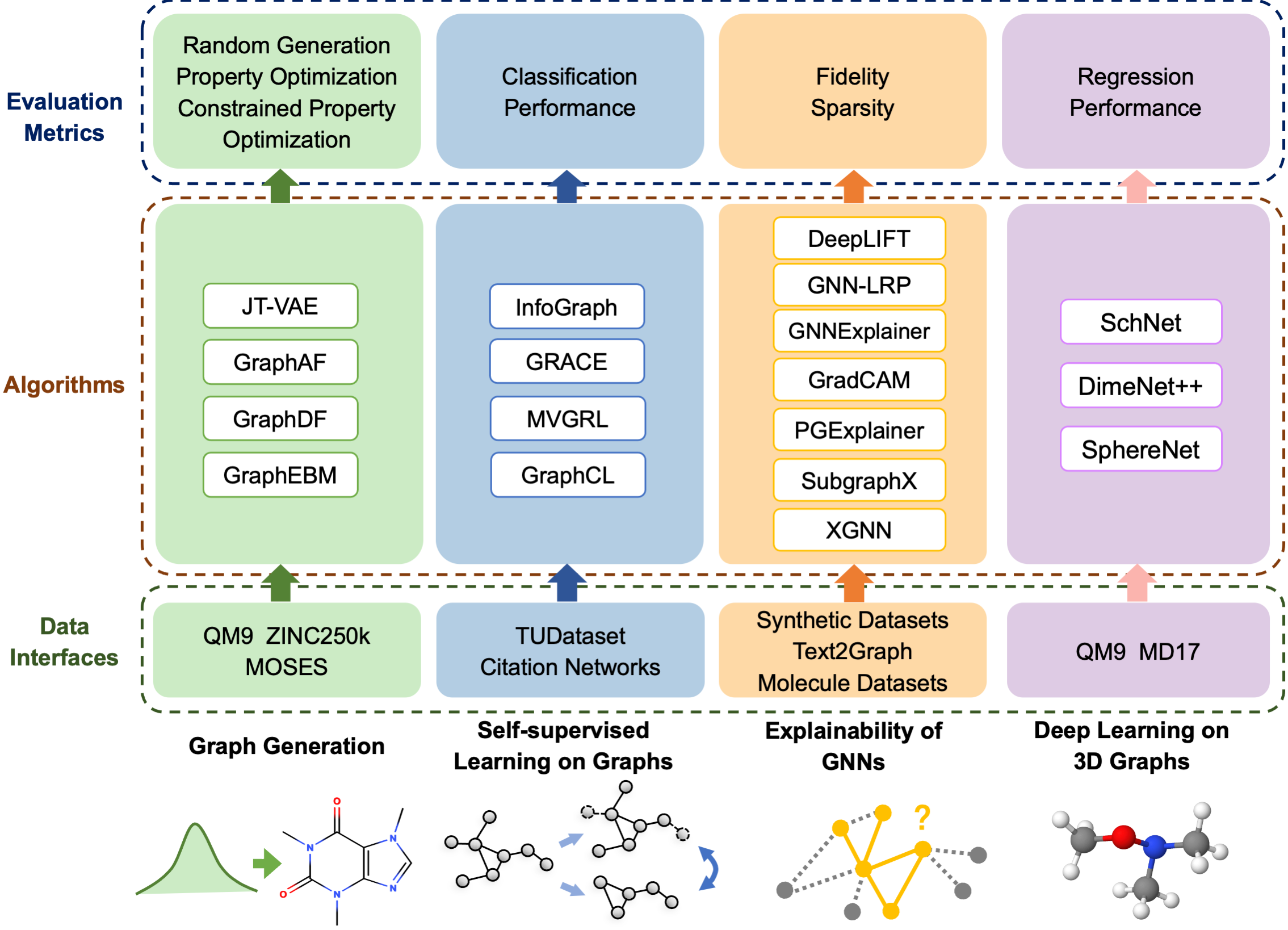}
    \end{center}\vspace{-0.4cm}
    \caption{A graphical overview of \textit{DIG: Dive into Graphs}.}
    \label{fig:overview}
\end{figure}

Our \textit{DIG} is based on Python and PyTorch~\citep{paszke2017automatic}. For some implementations, we also use PyG~\citep{Fey/Lenssen/2019} and RDKit~\citep{landrum2006rdkit} for basic operations on graphs and molecules. \textit{DIG} currently considers $4$ directions and contains $18$ algorithms. Note that more interesting directions and algorithms can be easily incorporated into \textit{DIG} based on the unified and extensible implementations. An overview of the \textit{DIG} library is illustrated in Figure~\ref{fig:overview}. We introduce the main implementations as follows.

% For each topic, introduce the problem and implemented algorithms, data interface, and evaluation metrics.

\textbf{Graph generation.} Given a set of graphs, graph generation algorithms aim at generating novel graphs~\citep{guo2020systematic,faez2020deep}. Graph generation is potentially useful for molecule discovery. Hence, we mainly consider algorithms that can generate molecular graphs. We include the following advanced algorithms: JT-VAE~\citep{jin2018junction}, GraphAF~\citep{shi2019graphaf}, GraphDF~\citep{luo2021graphdf}, and GraphEBM~\citep{liu2021graphebm}. We implement data interfaces for widely used datasets. These are QM9~\citep{ramakrishnan2014quantum}, ZINC250k~\citep{irwin2012zinc}, and MOSES~\citep{10.3389/fphar.2020.565644}. Metrics for evaluating random generation, property optimization, and constrained property optimization are also implemented as APIs.

\textbf{Self-supervised learning on graphs.} Self-supervised learning can help to obtain expressive representations by leveraging specified pretext tasks and has been recently extended to graph domain~\citep{jin2020self,xie2021self}. We incorporate InfoGraph~\citep{sun2019infograph}, GRACE~\citep{zhu2020deep}, MVGRL~\citep{hassani2020contrastive}, and GraphCL~\citep{You2020GraphCL} in \textit{DIG}. We provide the data interfaces of TUDataset (\emph{i.e.}, NCI1, PROTEINS, \emph{etc.})~\citep{Morris+2020} for graph-level classification tasks, and citation networks (\emph{i.e.}, Cora, CiteSeer, and PubMed)~\citep{yang2016revisiting} for node-level classification tasks. Standard metrics are also realized to evaluate the classification performance.

\textbf{Explainability of graph neural networks.} Since graph neural networks have been increasingly deployed in our real-world applications, it is critical to develop explanation techniques for better understanding of models~\citep{yuan2020explainability}. We include the following algorithms: GNNExplainer~\citep{ying2019gnnexplainer}, PGExplainer~\citep{luo2020parameterized}, DeepLIFT~\citep{shrikumar2017learning}, GNN-LRP~\citep{schnake2020higher}, Grad-CAM~\citep{pope2019explainability}, SubgraphX~\citep{yuan2021explainability}, and XGNN~\citep{yuan2020xgnn}. For data interfaces, we consider the widely used synthetic datasets (\emph{i.e.}, BA-shapes, BA-Community, \emph{etc.})~\citep{ying2019gnnexplainer,luo2020parameterized} and molecule datasets (\emph{i.e.}, BBBP, Tox21, \emph{etc.})~\citep{wu2018moleculenet}. In addition, we also build human-understandable graph datasets from text data and provide the corresponding data interfaces. Details of our proposed datasets (\emph{i.e.}, 
Graph-SST2, Graph-SST5, \emph{etc.}) are described by~\cite{yuan2020explainability}. Recently proposed metrics for explanation tasks, including Fidelity and Sparsity~\citep{pope2019explainability}, are implemented in our \textit{DIG}.

\textbf{Deep learning on 3D graphs.} 3D Graphs refer to graphs whose nodes are associated with 3D positions. For instance, in molecules, each atom has a relative 3D position. It is significant to investigate how to obtain expressive graph representations with such essential information. We consider three algorithms in the unified 3DGN framework~\citep{liu2021spherical}. These are SchNet~\citep{schutt2017schnet}, DimeNet++~\citep{klicpera_dimenet_2020,klicpera_dimenetpp_2020}, and SphereNet~\citep{liu2021spherical}. We implement data interfaces for two benchmark datasets: QM9~\citep{ramakrishnan2014quantum} and MD17~\citep{chmiela2017machine}. We apply mean absolute error (MAE), a standard metric for regression tasks, as the evaluation technique.

\section{Key Design Considerations}

% \textbf{Unified implementation.} As described in Section~\ref{sec:library} and illustrated in Figure~\ref{fig:overview}, we have unified implementations of data interfaces and evaluation metrics for each direction. This provides a standardized testbed for various algorithms in each direction. In addition, we provide unified implementations for different algorithms if they 
% enjoy non-trivial commonalities. To be specific, implementations of the three algorithms on 3D graphs can be unified using the 3DGN framework~\citep{liu2021spherical} with different internal functions. Also, many self-supervised learning algorithms on graphs can be viewed as contrastive models~\citep{xie2021self}. Hence, we provide unified objective functions for these algorithms.

In this section, we described the key design considerations of \textit{DIG}, including unified implementation, extensibility, and customization.

\textbf{Unified implementation.} As described in Section~\ref{sec:library} and illustrated in Figure~\ref{fig:overview}, we provide APIs of data interfaces, common algorithms, and evaluation metrics for each direction. This provides a standardized testbed for various algorithms in each direction. In addition, our implementations are unified for different algorithms if they 
enjoy non-trivial commonalities. To be specific, implementations of the three algorithms on 3D graphs can be unified using the 3DGN framework~\citep{liu2021spherical} with different internal functions. Also, many self-supervised learning algorithms on graphs can be viewed as contrastive models~\citep{xie2021self}. Hence, we provide unified objective functions for these algorithms.

\textbf{Extensibility and customization.} As a benefit of our unified implementations, it is easy to incorporate new datasets, algorithms, and evaluation metrics. Additionally, users can customize their own experiments on their new algorithms by flexibly choosing desired data interfaces and evaluation metrics. Therefore, our \textit{DIG} can serve as a platform for implementing and benchmarking algorithms in the covered directions.

\section{Quality Standards}

In the following, we evaluate our \textit{DIG} according to several quality standards of open source software.

\textbf{Code reliability and reproducibility.} The APIs of data interfaces and evaluation metrics in \textit{DIG} have been extensively tested by Travis CI, a continuous integration tool. In addition, for APIs corresponding to advanced algorithms, we provide the benchmark examples, which can reproduce the experimental results reported in the original papers within reasonable or negligible differences.

% mention various documentations: e.g., documentations for unified APIs, documentations for methods.

\textbf{Documentation.} \textit{DIG} has complete documentations online\footnote{\url{https://diveintographs.readthedocs.io}}, including the detailed descriptions of APIs and hands-on tutorials.

% Guidelines for contributing.

\textbf{Openness.} Contributions from the community are strongly welcome and encouraged. In our documented contribution guideline, we describe how to provide various types of contributions. The library is distributed under the GNU GPLv3 license.

\section{Conclusion and Outlook}

% Summary + Future paln
In this paper, we present \textit{DIG: Dive into Graphs} that contains unified and extensible implementations of data interfaces, common algorithms, and evaluation metrics for several significant research directions, including graph generation, self-supervised learning on graphs, explainability of graph neural networks, and deep learning on 3D graphs. We hope \textit{DIG} can enable researchers to easily implement and benchmark algorithms. In the future, we are interested in incorporating more emerging directions and advanced algorithms into \textit{DIG}.

% Acknowledgements should go at the end, before appendices and references

\acks{This work was supported in part by National Science Foundation grants IIS-2006861, IIS-1955189, IIS-1908220, IIS-1908198, DBI-2028361, and DBI-1922969.}


%\newpage

% \appendix
% \section*{Appendix A.}
% \label{app:theorem}

% % Note: in this sample, the section number is hard-coded in. Following
% % proper LaTeX conventions, it should properly be coded as a reference:

% %In this appendix we prove the following theorem from
% %Section~\ref{sec:textree-generalization}:

% In this appendix we prove the following theorem from
% Section~6.2:

% \vskip 0.2in
\bibliography{reference}

\begin{thebibliography}{56}
\providecommand{\natexlab}[1]{#1}
\providecommand{\url}[1]{\texttt{#1}}
\expandafter\ifx\csname urlstyle\endcsname\relax
  \providecommand{\doi}[1]{doi: #1}\else
  \providecommand{\doi}{doi: \begingroup \urlstyle{rm}\Url}\fi

\bibitem[Battaglia et~al.(2016)Battaglia, Pascanu, Lai, Rezende,
  et~al.]{battaglia2016interaction}
Peter Battaglia, Razvan Pascanu, Matthew Lai, Danilo~Jimenez Rezende, et~al.
\newblock Interaction networks for learning about objects, relations and
  physics.
\newblock In \emph{Advances in neural information processing systems}, pages
  4502--4510, 2016.

\bibitem[Battaglia et~al.(2018)Battaglia, Hamrick, Bapst, Sanchez-Gonzalez,
  Zambaldi, Malinowski, Tacchetti, Raposo, Santoro, Faulkner,
  et~al.]{battaglia2018relational}
Peter~W Battaglia, Jessica~B Hamrick, Victor Bapst, Alvaro Sanchez-Gonzalez,
  Vinicius Zambaldi, Mateusz Malinowski, Andrea Tacchetti, David Raposo, Adam
  Santoro, Ryan Faulkner, et~al.
\newblock Relational inductive biases, deep learning, and graph networks.
\newblock \emph{arXiv preprint arXiv:1806.01261}, 2018.

\bibitem[Bronstein et~al.(2017)Bronstein, Bruna, LeCun, Szlam, and
  Vandergheynst]{bronstein2017geometric}
Michael~M Bronstein, Joan Bruna, Yann LeCun, Arthur Szlam, and Pierre
  Vandergheynst.
\newblock Geometric deep learning: going beyond euclidean data.
\newblock \emph{IEEE Signal Processing Magazine}, 34\penalty0 (4):\penalty0
  18--42, 2017.

\bibitem[Cen et~al.(2021)Cen, Hou, Wang, Chen, Luo, Yao, Zeng, Guo, Zhang, Dai,
  Wang, Zhou, Yang, and Tang]{cen2021cogdl}
Yukuo Cen, Zhenyu Hou, Yan Wang, Qibin Chen, Yizhen Luo, Xingcheng Yao, Aohan
  Zeng, Shiguang Guo, Peng Zhang, Guohao Dai, Yu~Wang, Chang Zhou, Hongxia
  Yang, and Jie Tang.
\newblock {CogDL}: An extensive toolkit for deep learning on graphs.
\newblock \emph{arXiv preprint arXiv:2103.00959}, 2021.

\bibitem[Chmiela et~al.(2017)Chmiela, Tkatchenko, Sauceda, Poltavsky,
  Sch{\"u}tt, and M{\"u}ller]{chmiela2017machine}
Stefan Chmiela, Alexandre Tkatchenko, Huziel~E Sauceda, Igor Poltavsky,
  Kristof~T Sch{\"u}tt, and Klaus-Robert M{\"u}ller.
\newblock Machine learning of accurate energy-conserving molecular force
  fields.
\newblock \emph{Science advances}, 3\penalty0 (5):\penalty0 e1603015, 2017.

\bibitem[Data61(2018)]{StellarGraph}
CSIRO's Data61.
\newblock {StellarGraph} machine learning library.
\newblock \url{https://github.com/stellargraph/stellargraph}, 2018.

\bibitem[Faez et~al.(2020)Faez, Ommi, Baghshah, and Rabiee]{faez2020deep}
Faezeh Faez, Yassaman Ommi, Mahdieh~Soleymani Baghshah, and Hamid~R Rabiee.
\newblock Deep graph generators: A survey.
\newblock \emph{arXiv preprint arXiv:2012.15544}, 2020.

\bibitem[Fey and Lenssen(2019)]{Fey/Lenssen/2019}
Matthias Fey and Jan~E. Lenssen.
\newblock Fast graph representation learning with {PyTorch Geometric}.
\newblock In \emph{ICLR Workshop on Representation Learning on Graphs and
  Manifolds}, 2019.

\bibitem[Gilmer et~al.(2017)Gilmer, Schoenholz, Riley, Vinyals, and
  Dahl]{gilmer2017neural}
Justin Gilmer, Samuel~S Schoenholz, Patrick~F Riley, Oriol Vinyals, and
  George~E Dahl.
\newblock Neural message passing for quantum chemistry.
\newblock In \emph{Proceedings of the 34th international conference on machine
  learning}, pages 1263--1272, 2017.

\bibitem[Grattarola and Alippi(2020)]{grattarola2020graph}
Daniele Grattarola and Cesare Alippi.
\newblock Graph neural networks in tensorflow and keras with spektral.
\newblock \emph{arXiv preprint arXiv:2006.12138}, 2020.

\bibitem[Guo and Zhao(2020)]{guo2020systematic}
Xiaojie Guo and Liang Zhao.
\newblock A systematic survey on deep generative models for graph generation.
\newblock \emph{arXiv preprint arXiv:2007.06686}, 2020.

\bibitem[Hamilton et~al.(2017{\natexlab{a}})Hamilton, Ying, and
  Leskovec]{hamilton2017inductive}
Will Hamilton, Zhitao Ying, and Jure Leskovec.
\newblock Inductive representation learning on large graphs.
\newblock In \emph{Advances in Neural Information Processing Systems}, pages
  1024--1034, 2017{\natexlab{a}}.

\bibitem[Hamilton(2020)]{hamilton2020graph}
William~L Hamilton.
\newblock Graph representation learning.
\newblock \emph{Synthesis Lectures on Artifical Intelligence and Machine
  Learning}, 14\penalty0 (3):\penalty0 1--159, 2020.

\bibitem[Hamilton et~al.(2017{\natexlab{b}})Hamilton, Ying, and
  Leskovec]{grl_ideb17}
William~L. Hamilton, Rex Ying, and Jure Leskovec.
\newblock Representation learning on graphs: Methods and applications.
\newblock \emph{{IEEE} Data Eng. Bull.}, 40\penalty0 (3):\penalty0 52--74,
  2017{\natexlab{b}}.

\bibitem[Hassani and Khasahmadi(2020)]{hassani2020contrastive}
Kaveh Hassani and Amir~Hosein Khasahmadi.
\newblock Contrastive multi-view representation learning on graphs.
\newblock In \emph{International Conference on Machine Learning}, pages
  4116--4126. PMLR, 2020.

\bibitem[Hu et~al.(2021)Hu, Qian, Fang, Wang, Zhao, Zhang, and
  Xu]{hu2021efficient}
Jun Hu, Shengsheng Qian, Quan Fang, Youze Wang, Quan Zhao, Huaiwen Zhang, and
  Changsheng Xu.
\newblock Efficient graph deep learning in tensorflow with tf\_geometric, 2021.

\bibitem[Hu et~al.(2020)Hu, Fey, Zitnik, Dong, Ren, Liu, Catasta, and
  Leskovec]{hu2020ogb}
Weihua Hu, Matthias Fey, Marinka Zitnik, Yuxiao Dong, Hongyu Ren, Bowen Liu,
  Michele Catasta, and Jure Leskovec.
\newblock Open graph benchmark: Datasets for machine learning on graphs.
\newblock \emph{arXiv preprint arXiv:2005.00687}, 2020.

\bibitem[Irwin et~al.(2012)Irwin, Sterling, Mysinger, Bolstad, and
  Coleman]{irwin2012zinc}
John~J Irwin, Teague Sterling, Michael~M Mysinger, Erin~S Bolstad, and Ryan~G
  Coleman.
\newblock {ZINC}: a free tool to discover chemistry for biology.
\newblock \emph{Journal of chemical information and modeling}, 52\penalty0
  (7):\penalty0 1757--1768, 2012.

\bibitem[Jin et~al.(2020)Jin, Derr, Liu, Wang, Wang, Liu, and
  Tang]{jin2020self}
Wei Jin, Tyler Derr, Haochen Liu, Yiqi Wang, Suhang Wang, Zitao Liu, and
  Jiliang Tang.
\newblock Self-supervised learning on graphs: Deep insights and new direction.
\newblock \emph{arXiv preprint arXiv:2006.10141}, 2020.

\bibitem[Jin et~al.(2018)Jin, Barzilay, and Jaakkola]{jin2018junction}
Wengong Jin, Regina Barzilay, and Tommi Jaakkola.
\newblock Junction tree variational autoencoder for molecular graph generation.
\newblock In \emph{International Conference on Machine Learning}, pages
  2323--2332, 2018.

\bibitem[Kipf and Welling(2017)]{kipf2016semi}
Thomas~N Kipf and Max Welling.
\newblock Semi-supervised classification with graph convolutional networks.
\newblock In \emph{International Conference on Learning Representations}, 2017.

\bibitem[Klicpera et~al.(2020{\natexlab{a}})Klicpera, Giri, Margraf, and
  G{\"u}nnemann]{klicpera_dimenetpp_2020}
Johannes Klicpera, Shankari Giri, Johannes~T. Margraf, and Stephan
  G{\"u}nnemann.
\newblock Fast and uncertainty-aware directional message passing for
  non-equilibrium molecules.
\newblock In \emph{NeurIPS-W}, 2020{\natexlab{a}}.

\bibitem[Klicpera et~al.(2020{\natexlab{b}})Klicpera, Gro{\ss}, and
  G{\"u}nnemann]{klicpera_dimenet_2020}
Johannes Klicpera, Janek Gro{\ss}, and Stephan G{\"u}nnemann.
\newblock Directional message passing for molecular graphs.
\newblock In \emph{International Conference on Learning Representations
  (ICLR)}, 2020{\natexlab{b}}.

\bibitem[Landrum et~al.(2006)]{landrum2006rdkit}
Greg Landrum et~al.
\newblock {RDKit}: Open-source cheminformatics.
\newblock 2006.

\bibitem[Li et~al.(2021)Li, Xu, Chen, Zheng, and Liu]{li2021graphgallery}
Jintang Li, Kun Xu, Liang Chen, Zibin Zheng, and Xiao Liu.
\newblock Graphgallery: A platform for fast benchmarking and easy development
  of graph neural networks based intelligent software.
\newblock \emph{arXiv preprint arXiv:2102.07933}, 2021.

\bibitem[Liu et~al.(2021{\natexlab{a}})Liu, Yan, Oztekin, and
  Ji]{liu2021graphebm}
Meng Liu, Keqiang Yan, Bora Oztekin, and Shuiwang Ji.
\newblock {GraphEBM}: Molecular graph generation with energy-based models.
\newblock \emph{arXiv preprint arXiv:2102.00546}, 2021{\natexlab{a}}.

\bibitem[Liu et~al.(2021{\natexlab{b}})Liu, Wang, Liu, Zhang, Oztekin, and
  Ji]{liu2021spherical}
Yi~Liu, Limei Wang, Meng Liu, Xuan Zhang, Bora Oztekin, and Shuiwang Ji.
\newblock Spherical message passing for 3d graph networks.
\newblock \emph{arXiv preprint arXiv:2102.05013}, 2021{\natexlab{b}}.

\bibitem[Luo et~al.(2020)Luo, Cheng, Xu, Yu, Zong, Chen, and
  Zhang]{luo2020parameterized}
Dongsheng Luo, Wei Cheng, Dongkuan Xu, Wenchao Yu, Bo~Zong, Haifeng Chen, and
  Xiang Zhang.
\newblock Parameterized explainer for graph neural network.
\newblock \emph{Advances in Neural Information Processing Systems}, 33, 2020.

\bibitem[Luo et~al.(2021)Luo, Yan, and Ji]{luo2021graphdf}
Youzhi Luo, Keqiang Yan, and Shuiwang Ji.
\newblock {GraphDF}: A discrete flow model for molecular graph generation.
\newblock In \emph{International Conference on Machine Learning}, pages
  7192--7203, 2021.

\bibitem[Ma and Tang(2020)]{ma2020deep}
Yao Ma and Jiliang Tang.
\newblock \emph{Deep Learning on Graphs}.
\newblock Cambridge University Press, 2020.

\bibitem[Morris et~al.(2020)Morris, Kriege, Bause, Kersting, Mutzel, and
  Neumann]{Morris+2020}
Christopher Morris, Nils~M. Kriege, Franka Bause, Kristian Kersting, Petra
  Mutzel, and Marion Neumann.
\newblock {TUDataset}: A collection of benchmark datasets for learning with
  graphs.
\newblock In \emph{ICML 2020 Workshop on Graph Representation Learning and
  Beyond (GRL+ 2020)}, 2020.
\newblock URL \url{www.graphlearning.io}.

\bibitem[Paszke et~al.(2017)Paszke, Gross, Chintala, Chanan, Yang, DeVito, Lin,
  Desmaison, Antiga, and Lerer]{paszke2017automatic}
Adam Paszke, Sam Gross, Soumith Chintala, Gregory Chanan, Edward Yang, Zachary
  DeVito, Zeming Lin, Alban Desmaison, Luca Antiga, and Adam Lerer.
\newblock Automatic differentiation in pytorch.
\newblock 2017.

\bibitem[Polykovskiy et~al.(2020)Polykovskiy, Zhebrak, Sanchez-Lengeling,
  Golovanov, Tatanov, Belyaev, Kurbanov, Artamonov, Aladinskiy, Veselov,
  Kadurin, Johansson, Chen, Nikolenko, Aspuru-Guzik, and
  Zhavoronkov]{10.3389/fphar.2020.565644}
Daniil Polykovskiy, Alexander Zhebrak, Benjamin Sanchez-Lengeling, Sergey
  Golovanov, Oktai Tatanov, Stanislav Belyaev, Rauf Kurbanov, Aleksey
  Artamonov, Vladimir Aladinskiy, Mark Veselov, Artur Kadurin, Simon Johansson,
  Hongming Chen, Sergey Nikolenko, Alan Aspuru-Guzik, and Alex Zhavoronkov.
\newblock {M}olecular {S}ets ({MOSES}): {A} {B}enchmarking {P}latform for
  {M}olecular {G}eneration {M}odels.
\newblock \emph{Frontiers in Pharmacology}, 2020.

\bibitem[Pope et~al.(2019)Pope, Kolouri, Rostami, Martin, and
  Hoffmann]{pope2019explainability}
Phillip~E Pope, Soheil Kolouri, Mohammad Rostami, Charles~E Martin, and Heiko
  Hoffmann.
\newblock Explainability methods for graph convolutional neural networks.
\newblock In \emph{Proceedings of the IEEE/CVF Conference on Computer Vision
  and Pattern Recognition}, pages 10772--10781, 2019.

\bibitem[Ramakrishnan et~al.(2014)Ramakrishnan, Dral, Rupp, and
  Von~Lilienfeld]{ramakrishnan2014quantum}
Raghunathan Ramakrishnan, Pavlo~O Dral, Matthias Rupp, and O~Anatole
  Von~Lilienfeld.
\newblock Quantum chemistry structures and properties of 134 kilo molecules.
\newblock \emph{Scientific data}, 1\penalty0 (1):\penalty0 1--7, 2014.

\bibitem[Sanchez-Gonzalez et~al.(2020)Sanchez-Gonzalez, Godwin, Pfaff, Ying,
  Leskovec, and Battaglia]{sanchez2020learning}
Alvaro Sanchez-Gonzalez, Jonathan Godwin, Tobias Pfaff, Rex Ying, Jure
  Leskovec, and Peter Battaglia.
\newblock Learning to simulate complex physics with graph networks.
\newblock In \emph{International Conference on Machine Learning}, pages
  8459--8468. PMLR, 2020.

\bibitem[Schnake et~al.(2020)Schnake, Eberle, Lederer, Nakajima, Sch{\"u}tt,
  M{\"u}ller, and Montavon]{schnake2020higher}
T~Schnake, O~Eberle, J~Lederer, S~Nakajima, KT~Sch{\"u}tt, KR~M{\"u}ller, and
  G~Montavon.
\newblock Higher-order explanations of graph neural networks via relevant
  walks.
\newblock \emph{arXiv: 2006.03589}, 2020.

\bibitem[Sch{\"u}tt et~al.(2017)Sch{\"u}tt, Kindermans, Sauceda, Chmiela,
  Tkatchenko, and M{\"u}ller]{schutt2017schnet}
Kristof~T Sch{\"u}tt, PJ~Kindermans, Huziel~E Sauceda, Stefan Chmiela,
  Alexandre Tkatchenko, and Klaus~R M{\"u}ller.
\newblock {SchNet}: A continuous-filter convolutional neural network for
  modeling quantum interactions.
\newblock In \emph{Advances in Neural Information Processing Systems}, pages
  1--11, 2017.

\bibitem[Shi et~al.(2019)Shi, Xu, Zhu, Zhang, Zhang, and Tang]{shi2019graphaf}
Chence Shi, Minkai Xu, Zhaocheng Zhu, Weinan Zhang, Ming Zhang, and Jian Tang.
\newblock {GraphAF}: a flow-based autoregressive model for molecular graph
  generation.
\newblock In \emph{International Conference on Learning Representations}, 2019.

\bibitem[Shrikumar et~al.(2017)Shrikumar, Greenside, and
  Kundaje]{shrikumar2017learning}
Avanti Shrikumar, Peyton Greenside, and Anshul Kundaje.
\newblock Learning important features through propagating activation
  differences.
\newblock In \emph{International Conference on Machine Learning}, pages
  3145--3153. PMLR, 2017.

\bibitem[Stokes et~al.(2020)Stokes, Yang, Swanson, Jin, Cubillos-Ruiz, Donghia,
  MacNair, French, Carfrae, Bloom-Ackerman, et~al.]{stokes2020deep}
Jonathan~M Stokes, Kevin Yang, Kyle Swanson, Wengong Jin, Andres Cubillos-Ruiz,
  Nina~M Donghia, Craig~R MacNair, Shawn French, Lindsey~A Carfrae, Zohar
  Bloom-Ackerman, et~al.
\newblock A deep learning approach to antibiotic discovery.
\newblock \emph{Cell}, 180\penalty0 (4):\penalty0 688--702, 2020.

\bibitem[Sun et~al.(2020)Sun, Hoffmann, Verma, and Tang]{sun2019infograph}
Fan-Yun Sun, Jordan Hoffmann, Vikas Verma, and Jian Tang.
\newblock {InfoGraph}: Unsupervised and semi-supervised graph-level
  representation learning via mutual information maximization.
\newblock In \emph{International Conference on Learning Representations}, 2020.

\bibitem[Veli{\v{c}}kovi{\'c} et~al.(2018)Veli{\v{c}}kovi{\'c}, Cucurull,
  Casanova, Romero, Lio, and Bengio]{velivckovic2017graph}
Petar Veli{\v{c}}kovi{\'c}, Guillem Cucurull, Arantxa Casanova, Adriana Romero,
  Pietro Lio, and Yoshua Bengio.
\newblock Graph attention networks.
\newblock In \emph{International Conference on Learning Representation}, 2018.

\bibitem[Wang et~al.(2019)Wang, Zheng, Ye, Gan, Li, Song, Zhou, Ma, Yu, Gai,
  Xiao, He, Karypis, Li, and Zhang]{wang2019dgl}
Minjie Wang, Da~Zheng, Zihao Ye, Quan Gan, Mufei Li, Xiang Song, Jinjing Zhou,
  Chao Ma, Lingfan Yu, Yu~Gai, Tianjun Xiao, Tong He, George Karypis, Jinyang
  Li, and Zheng Zhang.
\newblock Deep graph library: A graph-centric, highly-performant package for
  graph neural networks.
\newblock \emph{arXiv preprint arXiv:1909.01315}, 2019.

\bibitem[Wang et~al.(2020)Wang, Liu, Luo, Xu, Xie, Wang, Cai, and
  Ji]{wang2020advanced}
Zhengyang Wang, Meng Liu, Youzhi Luo, Zhao Xu, Yaochen Xie, Limei Wang, Lei
  Cai, and Shuiwang Ji.
\newblock Advanced graph and sequence neural networks for molecular property
  prediction and drug discovery.
\newblock \emph{arXiv preprint arXiv:2012.01981}, 2020.

\bibitem[Wu et~al.(2018)Wu, Ramsundar, Feinberg, Gomes, Geniesse, Pappu,
  Leswing, and Pande]{wu2018moleculenet}
Zhenqin Wu, Bharath Ramsundar, Evan~N Feinberg, Joseph Gomes, Caleb Geniesse,
  Aneesh~S Pappu, Karl Leswing, and Vijay Pande.
\newblock {MoleculeNet}: a benchmark for molecular machine learning.
\newblock \emph{Chemical science}, 9\penalty0 (2):\penalty0 513--530, 2018.

\bibitem[Wu et~al.(2020)Wu, Pan, Chen, Long, Zhang, and
  Philip]{wu2020comprehensive}
Zonghan Wu, Shirui Pan, Fengwen Chen, Guodong Long, Chengqi Zhang, and S~Yu
  Philip.
\newblock A comprehensive survey on graph neural networks.
\newblock \emph{IEEE transactions on neural networks and learning systems},
  2020.

\bibitem[Xie et~al.(2021)Xie, Xu, Wang, and Ji]{xie2021self}
Yaochen Xie, Zhao Xu, Zhengyang Wang, and Shuiwang Ji.
\newblock Self-supervised learning of graph neural networks: A unified review.
\newblock \emph{arXiv preprint arXiv:2102.10757}, 2021.

\bibitem[Yang et~al.(2016)Yang, Cohen, and Salakhudinov]{yang2016revisiting}
Zhilin Yang, William Cohen, and Ruslan Salakhudinov.
\newblock Revisiting semi-supervised learning with graph embeddings.
\newblock In \emph{International conference on machine learning}, pages 40--48.
  PMLR, 2016.

\bibitem[Ying et~al.(2019)Ying, Bourgeois, You, Zitnik, and
  Leskovec]{ying2019gnnexplainer}
Rex Ying, Dylan Bourgeois, Jiaxuan You, Marinka Zitnik, and Jure Leskovec.
\newblock {GNNExplainer}: Generating explanations for graph neural networks.
\newblock \emph{Advances in neural information processing systems},
  32:\penalty0 9240, 2019.

\bibitem[You et~al.(2020)You, Chen, Sui, Chen, Wang, and Shen]{You2020GraphCL}
Yuning You, Tianlong Chen, Yongduo Sui, Ting Chen, Zhangyang Wang, and Yang
  Shen.
\newblock Graph contrastive learning with augmentations.
\newblock In \emph{Advances in Neural Information Processing Systems},
  volume~33, pages 5812--5823, 2020.

\bibitem[Yuan et~al.(2020{\natexlab{a}})Yuan, Tang, Hu, and Ji]{yuan2020xgnn}
Hao Yuan, Jiliang Tang, Xia Hu, and Shuiwang Ji.
\newblock {XGNN}: Towards model-level explanations of graph neural networks.
\newblock In \emph{Proceedings of the 26th ACM SIGKDD International Conference
  on Knowledge Discovery \& Data Mining}, pages 430--438, 2020{\natexlab{a}}.

\bibitem[Yuan et~al.(2020{\natexlab{b}})Yuan, Yu, Gui, and
  Ji]{yuan2020explainability}
Hao Yuan, Haiyang Yu, Shurui Gui, and Shuiwang Ji.
\newblock Explainability in graph neural networks: A taxonomic survey.
\newblock \emph{arXiv preprint arXiv:2012.15445}, 2020{\natexlab{b}}.

\bibitem[Yuan et~al.(2021)Yuan, Yu, Wang, Li, and Ji]{yuan2021explainability}
Hao Yuan, Haiyang Yu, Jie Wang, Kang Li, and Shuiwang Ji.
\newblock On explainability of graph neural networks via subgraph explorations.
\newblock In \emph{International Conference on Machine Learning}, pages
  12241--12252, 2021.

\bibitem[Zhou et~al.(2018)Zhou, Cui, Zhang, Yang, Liu, Wang, Li, and
  Sun]{zhou2018graph}
Jie Zhou, Ganqu Cui, Zhengyan Zhang, Cheng Yang, Zhiyuan Liu, Lifeng Wang,
  Changcheng Li, and Maosong Sun.
\newblock Graph neural networks: A review of methods and applications.
\newblock \emph{arXiv preprint arXiv:1812.08434}, 2018.

\bibitem[Zhu et~al.(2020)Zhu, Xu, Yu, Liu, Wu, and Wang]{zhu2020deep}
Yanqiao Zhu, Yichen Xu, Feng Yu, Qiang Liu, Shu Wu, and Liang Wang.
\newblock Deep graph contrastive representation learning.
\newblock \emph{arXiv preprint arXiv:2006.04131}, 2020.

\end{thebibliography}

\end{document}